\documentclass[journal]{IEEEtran}
\IEEEoverridecommandlockouts    

\usepackage{graphics}
\usepackage{epsfig}
\usepackage{times}
\usepackage{amsmath}
\usepackage{amssymb}
\usepackage[T1]{fontenc}
\usepackage{cite}
\usepackage{multirow,color}
\usepackage{booktabs}
\usepackage{makecell}
\usepackage[hidelinks]{hyperref}
\usepackage{threeparttable}
\usepackage{algorithm}
\usepackage{algorithmic}

\title{\LARGE \bf
TransFusion: A Practical and Effective Transformer-based Diffusion Model for 3D Human Motion Prediction
}

\author{Sibo Tian$^{1}$, Minghui Zheng$^{1,*}$, and Xiao Liang$^{2,*}$
\thanks{This work was supported by the USA National Science Foundation under Grant No. 2026533. The authors confirm that all human/animal subject research procedures and protocols are exempt from review board approval.}
\thanks{$^{1}$ Sibo Tian and Minghui Zheng are with the Department of Mechanical and Aerospace Engineering, University at Buffalo, Buffalo, NY 14260, USA.
        {\tt\small Emails: {sibotian, mhzheng}@buffalo.edu.}}
\thanks{$^{2}$ Xiao Liang is with the Department of Civil, Structural and Environmental Engineering, University at Buffalo, Buffalo, NY 14260, USA.
        {\tt\small Email: liangx@buffalo.edu.}}
\thanks{$^{*}$ Corresponding Authors.}
}

\begin{document}

\maketitle
\thispagestyle{plain}
\pagestyle{plain}

\begin{abstract}
Predicting human motion plays a crucial role in ensuring a safe and effective human-robot close collaboration in intelligent remanufacturing systems of the future. Existing works can be categorized into two groups: those focusing on accuracy, predicting a single future motion, and those generating diverse predictions based on observations. The former group fails to address the uncertainty and multi-modal nature of human motion, while the latter group often produces motion sequences that deviate too far from the ground truth or become unrealistic within historical contexts. To tackle these issues, we propose TransFusion, an innovative and practical diffusion-based model for 3D human motion prediction which can generate samples that are more likely to happen while maintaining a certain level of diversity. Our model leverages Transformer as the backbone with long skip connections between shallow and deep layers. Additionally, we employ the discrete cosine transform to model motion sequences in the frequency space, thereby improving performance. In contrast to prior diffusion-based models that utilize extra modules like cross-attention and adaptive layer normalization to condition the prediction on past observed motion, we treat all inputs, including conditions, as tokens to create a more lightweight model compared to existing approaches. Extensive experimental studies are conducted on benchmark datasets to validate the effectiveness of our human motion prediction model.
\end{abstract}

\begin{IEEEkeywords}
Human Motion Prediction, Deep Learning, Diffusion Models, Human-Robot Collaboration (HRC)
\end{IEEEkeywords}

\section{INTRODUCTION}
Human-robot collaboration (HRC) in the recycling of end-of-life electronic products has gained significant attention in recent years \cite{c1,c2,c3,c4}. Unlike traditional remanufacturing processes, where industrial robots and human workers perform separate tasks in isolation for safety purposes, HRC allows for the synergistic utilization of both human workers and robot agents during collaborative disassembly. Humans excel at handling uncertainty and making flexible decisions, while robots are efficient at performing repetitive, labor-intensive, and hazardous tasks. When humans and robots work in close proximity, such as performing alternating tasks or cooperating to complete a task together (e.g., handing over tools or disassembled parts), it is crucial for robots to understand their collaborator's behaviors to ensure safety and improve collaboration efficiency. Therefore, modeling human behavior and predicting human future motion are essential for achieving safe and seamless HRC. Many prior works \cite{c5,c5-1, c10,c11,c12,c13,c14,c15,c16,c17,c18,c19,c20,c21,c22,c23, c25, c26,c27,c28,c29,c30,c31,c32,c33,c34,c35,c40,c41,c42,c43} have explored these areas.

    \begin{figure}
      \centering
      \includegraphics[width=1.0\columnwidth]{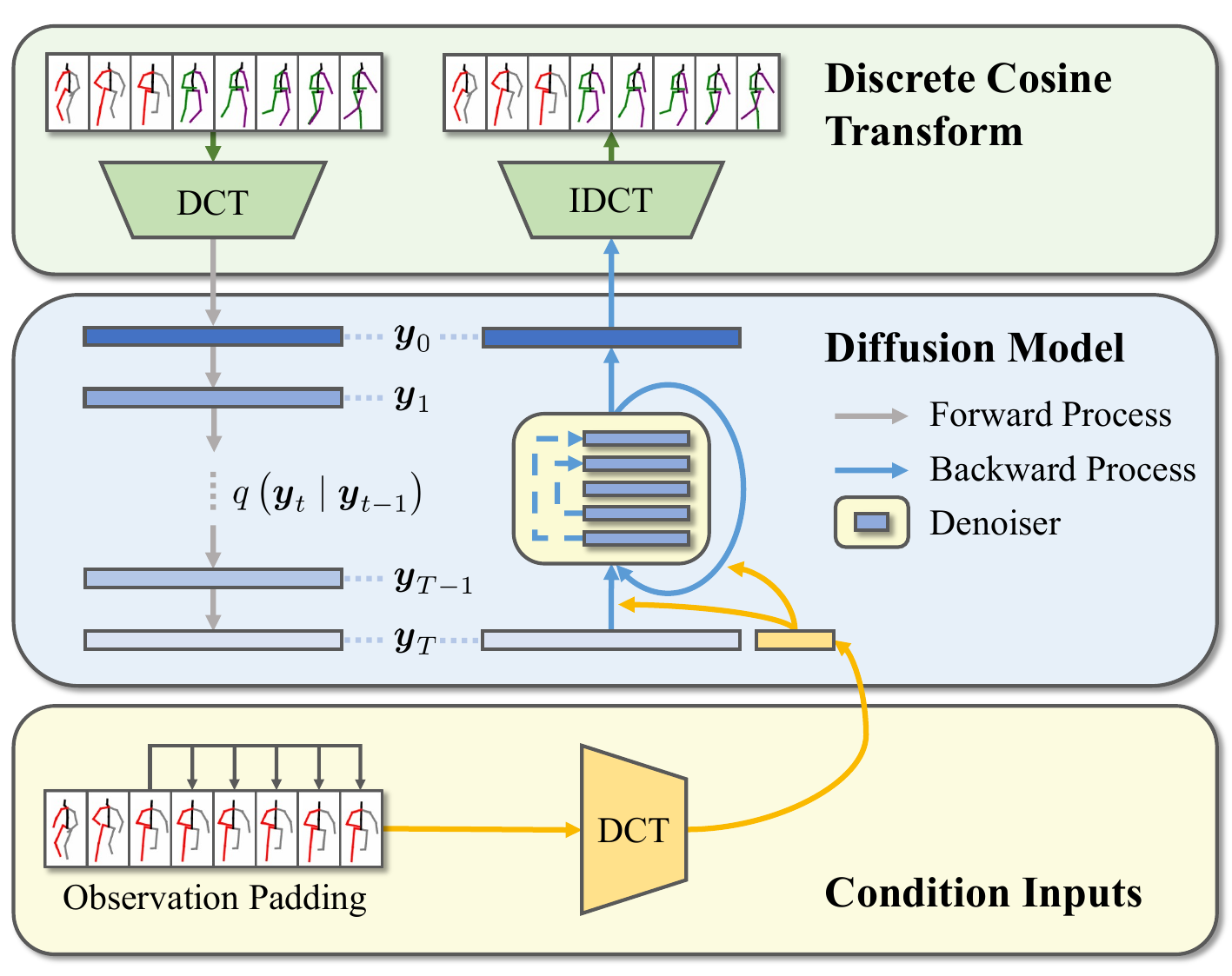}
        \vspace{-0.1 in}
      \caption{An overview of the proposed human motion prediction method. The proposed human motion prediction method consists of a diffusion process and a reverse process. In the diffusion process, the motion sequence is transformed to the frequency domain using DCT. The noise is then progressively incorporated into the data over $T$ diffusion steps, resulting in white noise representation. In the reverse process, the observation is padded to match the length of the motion sequence. After applying DCT, the condition inputs guide the denoiser in recovering the data from the pure noise representation. Finally, IDCT is applied to reconstruct the motion sequence from its frequency components.}
      \label{overview}
      \vspace{-0.2 in}
   \end{figure}

The human motion prediction problem involves predicting potential human poses based on a sequence of historical data. Neural networks have shown strong capacities in time series forecasting \cite{c6}, and the availability of large-scale human motion datasets like Human3.6M \cite{c7}, HumanEva-I \cite{c8}, and AMASS \cite{c9} has enabled exploration of neural network-based human motion prediction. Previous state-of-the-art works in neural network-based methods aimed to regress a single future sequence of human skeletal data based on observations. This regression problem was typically modeled with recurrent neural networks (RNNs) \cite{c10,c11,c12,c13} due to their ability to capture temporal correlation between sequential data. However, RNN-based approaches have notable drawbacks, including vanishing or exploding gradient problems \cite{c24}, error accumulation, and discontinuity between the first frame of the prediction and the last frame of the observation  \cite{c11}. To address these weaknesses, researchers have proposed using Transformers \cite{c14,c15,c16,c17} and graph convolutional networks (GCNs) \cite{c18,c19,c20,c21} in the human motion prediction task. These methods offer benefits such as learning the spatial-temporal dependencies of human motion data and reducing accumulated errors. Nevertheless, these models tend to be relatively complex. More recent works have returned to multi-layer perceptrons (MLPs) to dramatically reduce model complexity and computational cost \cite{c22,c23} while achieving state-of-the-art performance.

Considering the inherent uncertainty and multi-modality of human motion, it is crucial to predict the distribution of potential human motions rather than relying on a single deterministic output, especially for safety-critical applications like HRC. Recent research on stochastic human motion prediction has primarily focused on deep generative models. Previous works have utilized generative adversarial networks (GANs) \cite{c25,c26} and variational autoencoders (VAEs) \cite{c27,c28,c29,c30,c32,c33,c34,c35} to generate multiple samples of future motions based on a short observed sequence data. These works typically incorporate multiple loss constraints to ensure both the quality and diversity of generated samples, requiring careful parameter tuning to balance different loss functions. However, diversity-promoting techniques, such as diversity loss and diverse sampling, may lead to early deviations from the ground truth or sudden stagnation, resulting in unrealistic and implausible predictions given the context. Overly-diverse predictions can hinder downstream applications, such as motion planning for collaborative robot manipulators, as the excessive variety and out-of-context predictions may cover most of the shared working space when all predictions are considered valid. This leads to an overly conservative planned robot trajectory or even a failure to find a possible solution, contradicting the purpose of incorporating human motion prediction in HRC.

Recently, a new deep generative model called denoising diffusion probabilistic model (DDPM) \cite{c36} has shown significant progress in generative tasks, including image synthesis \cite{c37, c38} and image repainting \cite{c39}. Researchers have explored its potential for human motion prediction tasks \cite{c40,c41,c42,c43,c44}. In this work, we propose a new diffusion model that incorporates a simple yet powerful transformer-based denoising neural network. Specifically, unlike current state-of-the-art works that require additional modules to handle diffusion steps and conditions using cross-attention or adaptive normalization, we treat all the inputs, including diffusion steps and conditions, as tokens for the transformer. This reduces the complexity of the network. We also utilize long skip connections \cite{c45} to fuse the information from shallow and deep layers for improved training and employ squeeze and excitation (SE) blocks \cite{c46} to enhance prediction performance. In addition, instead of representing human motion in the time domain, we adopt the discrete cosine transform (DCT) and learn the model in the frequency domain. DCT helps reduce the dimension of motion sequences while preserving important details by eliminating high-frequency components, which are primarily noise. This approach is beneficial for predicting continuous motions as it extracts time properties from sequential data. Overall, the main contributions of this work can be summarized as follows:

\begin{itemize}
\item We provide a detailed review of all diffusion-based human motion prediction works known to us, which can serve as inspiration for future research in this area.
\item We propose a novel, practical, and effective diffusion-based model called TransFusion for 3D Human motion prediction. Compared with prior works, TransFusion is more lightweight and achieves state-of-the-art accuracy on the Human3.6M and HumanEva-I datasets.
\item We conduct comprehensive experiments and ablation studies on two benchmark datasets to validate the performance of the proposed human motion prediction approach.
\end{itemize}

The rest of the paper is organized as follows. Section II presents related works. Section III describes the methodology of our human motion prediction model. Section IV showcases the experimental studies of the proposed method. Finally, Section V concludes this work.

\section{RELATED WORK}

\subsection{Diffusion Models}
Diffusion models \cite{c36,c47}, considered a new and promising addition to deep generative model family, have gained attention for their ability to generate high-quality samples through a simple training procedure. These models consist of two key processes: the forward process and the backward process. Inspired by the second law of thermodynamics, the forward process introduces noise gradually to the original data over multiple diffusion steps. Eventually, after $T$ steps, the data will turn into pure noise. This diffusion process can be represented as a Markov chain:
$$q\left(\boldsymbol{x}_{1: T} \mid \boldsymbol{x}_0\right)=\prod_{t=1}^T q\left(\boldsymbol{x}_t \mid \boldsymbol{x}_{t-1}\right) \eqno{(1)}$$where $\boldsymbol{x}_0$ represents the original data and $\boldsymbol{x}_t$ is the perturbed data at diffusion step $t$. The transition of each step is represented as: $$q\left(\boldsymbol{x}_t \mid \boldsymbol{x}_{t-1}\right)=\mathcal{N}\left(\boldsymbol{x}_t \mid \sqrt{\alpha_t} \boldsymbol{x}_{t-1}, \beta_t \boldsymbol{I}\right) \eqno{(2)}$$where $\beta_t$ is the pre-defined noise schedule, and $\alpha_t = 1-\beta_t$. Note that in the forward process, $\boldsymbol{x}_t$ for an arbitrary step $t$ can be sampled directly using $\boldsymbol{x}_0$ in a closed form with the notation $\bar{\alpha}_t = \prod_{i=1}^t \alpha_i$:$$q\left(\boldsymbol{x}_t \mid \boldsymbol{x}_0\right)=\mathcal{N}\left(\boldsymbol{x}_t \mid \sqrt{\bar{\alpha}_t} \boldsymbol{x}_0,\left(1-\bar{\alpha}_t\right) \boldsymbol{I}\right). \eqno{(3)}$$

Regarding the backward process, a natural approach is to reverse the steps applied in the forward process, aiming to restore the clean data from pure noise. To achieve this goal, a denoising process is proposed to approximate the true backward transition $q\left(\boldsymbol{x}_{t-1} \mid \boldsymbol{x}_{t}\right)$ by learning a Gaussian model: $$p_{\theta}\left(\boldsymbol{x}_{t-1} \mid \boldsymbol{x}_t\right)=\mathcal{N}\left(\boldsymbol{x}_{t-1} \mid \boldsymbol{\mu}_{\theta} \left(\boldsymbol{x}_t, t\right),\boldsymbol{\Sigma}_\theta\left(\mathbf{x}_t, t\right)\right). \eqno{(4)}$$Instead of predicting $\boldsymbol{x}_{t-1}$ from $\boldsymbol{x}_{t}$ directly at each step $t$, DDPM proposes that predicting the injected noise generates better results, and the mean $\boldsymbol{\mu}_{\theta} \left(\boldsymbol{x}_t, t\right)$ can be represented as: $$\boldsymbol{\mu}_{\theta} \left(\boldsymbol{x}_t, t\right) = \frac{1}{\sqrt{\alpha_t}}\left(\boldsymbol{x}_t-\frac{\beta_t}{\sqrt{1-\bar{\alpha}_t}} \boldsymbol{\epsilon}_\theta\left(\boldsymbol{x}_t, t\right)\right). \eqno{(5)}$$$\boldsymbol{\epsilon}_\theta\left(\boldsymbol{x}_t, t\right)$ is the noise-predicting neural network with a simple loss function:$$\begin{aligned}\mathcal{L}(\theta) & = \mathbb{E}_{t, \boldsymbol{x}_0, \epsilon} \left\|\boldsymbol{\epsilon}-\boldsymbol{\epsilon}_\theta\left(\boldsymbol{x}_t, t\right)\right\|^2_2 \\ & =\mathbb{E}_{t, \boldsymbol{x}_0, \epsilon} \left\|\boldsymbol{\epsilon}-\boldsymbol{\epsilon}_\theta\left(\sqrt{\bar\alpha_t} \boldsymbol{x}_0+\sqrt{1-\bar\alpha_t} \boldsymbol{\epsilon}, t\right)\right\|^2_2 . \end{aligned} \eqno{(6)}$$where $\boldsymbol{\epsilon} \sim \mathcal{N}(\boldsymbol{0}, \boldsymbol{I})$. In practice, $t$ in the loss function is sampled uniformly within the range of diffusion steps during the training process. As for the covariance $\boldsymbol{\Sigma}_\theta\left(\boldsymbol{x}_t, t\right)$, DDPM sets it as untrained time-dependent constants for simplicity, i.e., $\boldsymbol{\Sigma}_\theta\left(\boldsymbol{x}_t, t\right)=\sigma_t^2 \boldsymbol{I}$ where $\sigma_t^2 = \frac{1-\bar{\alpha}_{t-1}}{1-\bar{\alpha}_t} \beta_t $. Finally, $\boldsymbol{x}_{t-1}$ can be sampled from $p_{\theta}\left(\boldsymbol{x}_{t-1} \mid \boldsymbol{x}_t\right)$ as below: $$\boldsymbol{x}_{t-1} = \frac{1}{\sqrt{\alpha_t}}\left(\boldsymbol{x}_t-\frac{\beta_t}{\sqrt{1-\bar{\alpha}_t}} \boldsymbol{\epsilon}_\theta\left(\boldsymbol{x}_t, t\right)\right) + \sigma_t \boldsymbol{z} \eqno{(7)}$$where $\boldsymbol{z} \sim \mathcal{N}(\boldsymbol{0}, \boldsymbol{I})$ is the Gaussian noise.

While DDPM excels at generating high-quality samples without complex adversarial training, it does have a drawback in terms of inference time. The process of simulating a Markov chain multiple times to generate samples from pure noise can be time-consuming. To address this limitation, the denoising diffusion implicit model (DDIM) \cite{c48} was introduced. DDIM proposes a non-Markovian diffusion process to accelerate the sampling process while maintaining the generation of high-quality samples. Notably, DDIM modifies the inference stage of DDPM without altering the training procedure. 

\subsection{Diffusion-based Human Motion Prediction}

While prior works in human motion prediction have achieved good performance in predicting a single future motion sequence \cite{c10,c11,c12,c13,c14,c15,c16,c17,c18,c19,c20,c21,c22,c23}, these deterministic methods are limited in their ability to model uncertainty and multi-modal distributions of human motions. To address this limitation, several stochastic approaches have been explored, utilizing generative models like GANs and VAEs \cite{c25,c26,c27,c28,c29,c30,c31,c32,c33,c34,c35} to sample multiple plausible predictions from a given observed sequence. However, these approaches often involve complex loss functions with multiple terms, requiring careful parameter design and sometimes multiple training stages or adversarial training.

The emergence of diffusion models has provided a new promising direction for predicting human motion while considering uncertainty. Researchers have begun exploring the use of diffusion models in this context \cite{c40,c41,c42,c43,c44}. For example, one study \cite{c44} proposed using spatial and temporal Transformers arranged in series or in parallel as the motion denoiser. The observed past motion sequence and perturbed future motion sequence are concatenated together and passed into the noise-predicting network. The diffusion step $t$ is injected by first projecting it to the vector space and then adding it directly to the input sequence. While the model does not achieve state-of-the-art performance, it demonstrated that diffusion models can strike a balance between diversity and accuracy, generating motion predictions that are contextually appropriate.

Two works, MotionDiff \cite{c40} and TCD \cite{c42}, also utilized the spatial-temporal Transformer in the diffusion model. MotionDiff \cite{c40} employs an encoder-decoder structure, where the past motion sequence is first processed by the encoder network and then concatenated with the diffusion time encoding of step $t$. This encoded information serves as a condition for motion generation and is used multiple times within a single denoising step through a module that combines linear transformations with gating and bias addition. For the decoder part, MotionDiff utilizes the spatial-temporal Transformer \cite{c49} to extract information from the noised future motion sequence sampled from the distribution shown in Eq. (3). The hidden vector, along with the condition, is then fed into another Transformer block to predict the noise. After obtaining the outputs from the diffusion model, MotionDiff uses a GCN to refine the results, making the approach more intricate and preventing it from being trained in an end-to-end manner. On the other hand, TCD \cite{c42} adopts a similar approach to condition the denoiser on the observations and diffusion step $t$, much like as \cite{c44} does. It concatenates the past motion with the noised future motion and directly adds the embedded vector of diffusion step. However, TCD takes a different perspective on the prediction task by breaking it into two parts: short-term and long-term predictions. Instead of generating the entire prediction sequence at once given a short observed human motion, the short-term diffusion block aims to predict the first few frames of the future motion sequence based on the observation. The long-term diffusion block then generates the remaining frames of the future motion using both the observation and the outputs of the short-term diffusion block as the new condition. Another contribution of TCD is its ability to handle imperfect observation by introducing noise to the missing elements in the past motion sequence. The authors trained different models corresponding to different data-missing situations to validate the effectiveness of their approach.

In \cite{c43}, an end-to-end diffusion model called HumanMAC was proposed, which solves the prediction problem from the perspective of masked completion. Specifically, the model is trained to generate the entire motion sequence, encompassing both the observed and future motion, starting from random noise. During the inference stage, the future motion is treated as a missing part within the complete sequence, as only the observation part is available. In each denoising step, the noisy known region is sampled from the observation, and the inpainted region is sampled from the output of the previous iteration. These two samples are combined through a mask operation before being passed to the denoiser. Moreover, HumanMAC represents the motion sequence in the frequency space using DCT, thereby reducing the computational cost by discarding high-frequency components. Additionally, adaptive normalization modules are introduced after the self-attention layer and the feed-forward network in the Transformer to guide the prediction using historical information and diffusion steps.

    \begin{figure*}
      \begin{center}
      \includegraphics[width=1.9\columnwidth]{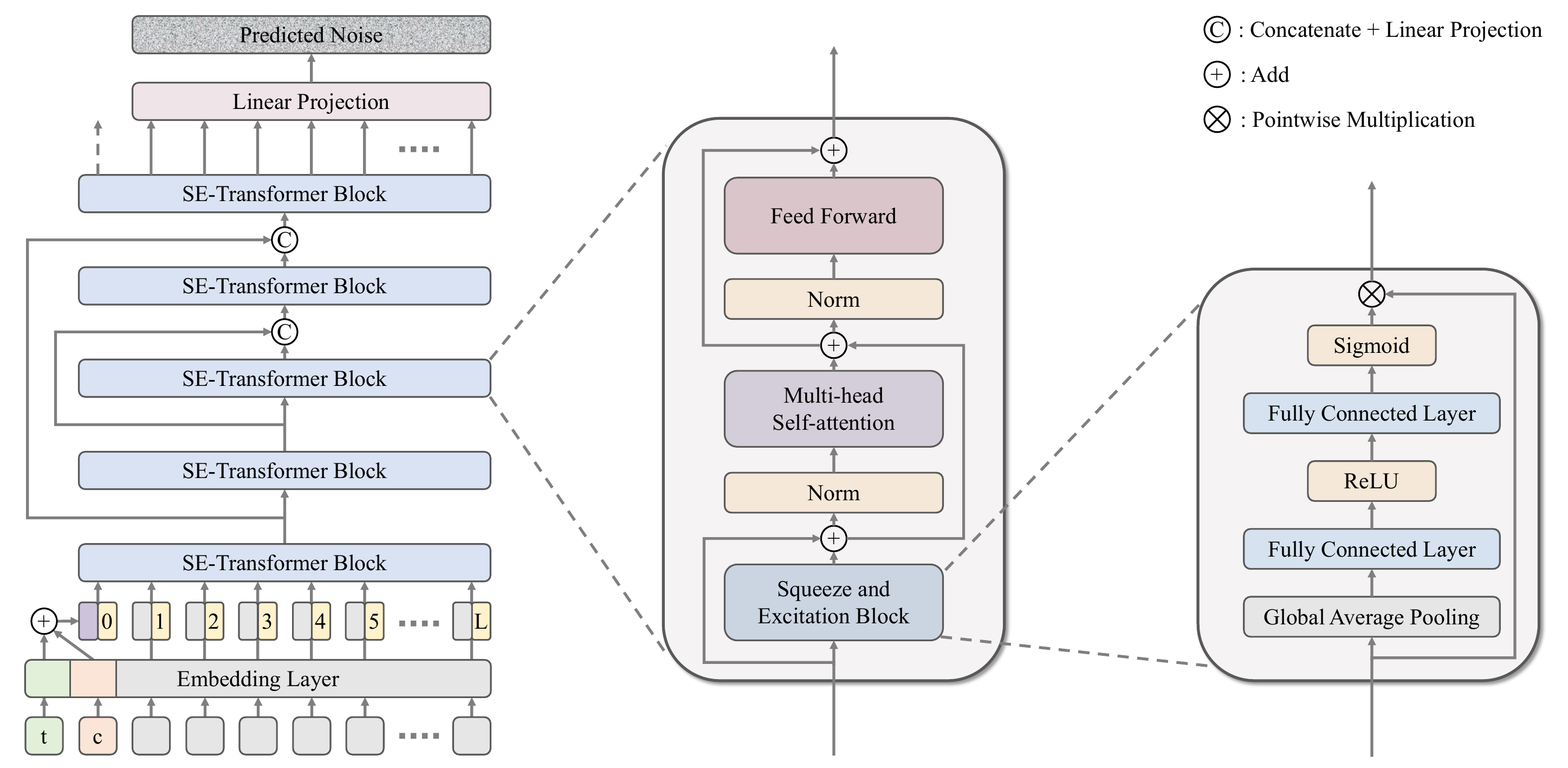}
      \caption{Architecture of the noise prediction network. The noise prediction network consists of several SE-Transformer blocks. At each diffusion step (t), and with the inclusion of historical information, the tokens are embedded and combined. Along with positional embeddings, these tokens are then processed through the SE-Transformer blocks. The final outputs from the last layer of SE-Transformer blocks are passed to a linear projection layer, which yields the predicted noise for the given input.}
      \label{architecture}
      \vspace{-0.1 in}
      \end{center}
   \end{figure*}
Unlike the aforementioned works, BeLFusion \cite{c41} takes a different approach by interpreting diversity from a behavioral perspective rather than focusing on skeleton joint dispersion. The diffusion model utilized in BeLFusion is based on the U-net with cross-attention \cite{c37}, which allows the model to sample behavior codes in the latent space. These behavior codes are then transferred to the ongoing motion through a behavior coupler, resulting in more realistic predictions. However, BeLFusion requires multiple training stages and complex adversarial training to disentangle behavior from pose and motion, which makes their model difficult to implement.

\section{METHOD}

In this section, we describe the methodology of TransFusion. First, we define the notations relevant to the human motion prediction task. Next, we tailor the DDPM to suit our specific scenario and introduce our TransFusion model. Lastly, we provide an overview of the training and inference procedure employed in TransFusion.

\subsection{Problem Definition and Notations}

We note the complete sequence of human motion as $\boldsymbol{x}=\left[\boldsymbol{q}^{(t-H)}, \ldots , \boldsymbol{q}^{(t-2)}, \boldsymbol{q}^{(t-1)}, \boldsymbol{q}^{(t)}, \boldsymbol{q}^{(t+1)}, \ldots , \boldsymbol{q}^{(t+F-1)}\right] \in \boldsymbol{R}^{(H+F) \times 3J}$, where $\boldsymbol{q}^{(t)} \in \boldsymbol{R}^{3J}$ is the Cartesian coordinates of human skeleton at the frame $t$, and $J$ is the number of human joints. The first $H$ frames of $\boldsymbol{x}$ correspond to the observation, denoted as $\boldsymbol{x}^O$, and the following $F$ frames represent the future motion $\boldsymbol{x}^P$ to be predicted. Given the observed human motion $\boldsymbol{x}^O$, the objective of human motion prediction consists in predicting the future motion sequence $\boldsymbol{x}^P$. We use $\boldsymbol{y}$ to represent the frequency components after the DCT operation.

\subsection{Transformer-based Diffusion Model (TransFusion)}

We propose a direct adaptation of the diffusion model to the human motion prediction problem. As shown in Fig. \ref{overview}, in the forward process, the motion sequence $\boldsymbol{x}$ is first projected to the frequency domain via the DCT operation:$$\boldsymbol{y} = \operatorname{DCT}(\boldsymbol{x}) = \boldsymbol{D}x \eqno{(8)}$$where $\boldsymbol{D} \in \boldsymbol{R}^{(H+F) \times (H+F)}$ is the DCT basis, and $\boldsymbol{y} \in \boldsymbol{R}^{(H+F) \times 3J}$ is DCT coefficients after transformation. The DCT is commonly used in human motion prediction \cite{c15, c16, c18, c19, c23, c34, c31, c43} due to its ability to encode the temporal nature of human motion. What's more, since the DCT operation is an orthogonal transform, we can always recover the original motion sequence from frequency coefficients by applying the inverse discrete cosine transform (IDCT):$$\boldsymbol{x} = \operatorname{IDCT}(\boldsymbol{y}) = \boldsymbol{D}^{\top} \boldsymbol{y}. \eqno{(9)}$$Considering that the most important and relevant information of human motion is concentrated in the lower frequency coefficients, and the higher frequency terms are mainly related to the noise, we simply keep the first $L$ rows of DCT basis and ignore the remaining rows to reduce the dimensionality of the data, i.e., $\boldsymbol{D} \in \boldsymbol{R}^{L \times (H+F)}$ and $\boldsymbol{y} \in \boldsymbol{R}^{L \times 3J}$. Then, the noisy DCT coefficients $\boldsymbol{y}_t$ at any diffusion step $t$ can be sampled by the reparameterization trick: $$\boldsymbol{y}_t = \sqrt{\bar\alpha_t} \boldsymbol{y}_0 + \sqrt{1 - \bar\alpha_t} \boldsymbol{\epsilon} \eqno{(10)}$$where $\boldsymbol{\epsilon} \sim \mathcal{N}(\boldsymbol{0}, \boldsymbol{I})$ and $\boldsymbol{y}_0$ equals to $\boldsymbol{y}$ in Eq. (8). 

Regarding noise prediction in the backward process for human motion sequences, we propose an alternative approach to the convolutional neural network-based U-net used in the original DDPM \cite{c36}. Our approach involves a Transformer-based network denoted as $\boldsymbol{\epsilon}_\theta$, and its architecture is given in Fig. \ref{architecture}. Specifically, in order to enable predictions conditioned on the observation, we leverage the conditional DDPM to guide the sample generation. The observation sequence $\boldsymbol{x}^O$ is first padded to match the length of the complete motion sequence, with the last frame of $\boldsymbol{x}^O$ extended accordingly. Subsequently, the padded sequence is processed through the DCT operator to obtain compact historical information. Different from prior works that employed extra modules like cross-attention and adaptive normalization to inject the observation and diffusion step, we calculate the condition by aggregating the encoded diffusion step and historical data. This condition, along with all the noisy DCT coefficients, are treated as tokens and fed into the denoiser network which comprises several Transformer layers with long skip connections between shallow and deep layers. The residual connection is achieved by concatenating two tensors and passing them through a linear projection layer. Each Transformer layer is composed of an SE block \cite{c46}, a self-attention module, and a feed-forward network. The SE block functions as an attention mechanism, but with significantly fewer parameters compared to the self-attention module. It consists of only two fully connected layers with a single pointwise multiplication. The introduction of the SE block allows for adaptive rescaling of each channel by modeling inter-dependencies among different channels, optimizing the Transformer encoder's learning process, and enhancing network performance.

Furthermore, in the inference stage, as we already have access to information about the historical motion, inspired by \cite{c43}, we propose the integration of noisy observation guidance at the beginning of each denoising step. This guidance involves several sequential operations. Firstly, we project the denoised DCT coefficients obtained from the previous denoising step and the noisy frequency coefficients acquired from the observation into the temporal domain using IDCT. Subsequently, these components are mixed together via a mask operation, where we define the mask as $\boldsymbol{M}=[1,1, \ldots, 1,0,0, \ldots 0]^{\top}$. The mask $\boldsymbol{M}$ consists of $H$ elements set to one, representing the noisy observation, and all other elements are set to zero, representing the denoised motion. To distinguish between samples from the last denoising step and the observation, we use $\boldsymbol{y}_t^D$ to denote the denoised samples and $\boldsymbol{y}_t^O$ to denote the observed samples. With these notations, the process of the noisy observation guidance can be summarized as follows:$$\begin{aligned}\boldsymbol{y}_{t}= \operatorname{DCT} & \left[\boldsymbol{M} \odot \operatorname{IDCT}\left(\boldsymbol{y}_{t}^{O}\right) + \right. \\ & \left. (\boldsymbol{1}-\boldsymbol{M}) \odot \operatorname{IDCT}\left(\boldsymbol{y}_{t}^{D} \right)\right]. \end{aligned} \eqno{(11)} $$

We present the workflow for model training and inference, as outlined in Algorithm \ref{alg1} and \ref{alg2}.

\begin{algorithm}
	\renewcommand{\algorithmicrequire}{\textbf{Input:}}
	\renewcommand{\algorithmicensure}{\textbf{Output:}}
	\caption{Training procedure of TransFusion}
	\label{alg1}
	\begin{algorithmic}[1]
            \REQUIRE complete motion sequence $\boldsymbol{x}$, diffusion steps $T$, denoiser $\epsilon_\theta$, maximum training epoch $E_{max}$.
		\FOR{$i = 0,1,\cdots,E_{max}$}
		\STATE $\boldsymbol{x}_0 \sim p\left(\boldsymbol{x}\right)$
		\STATE $\boldsymbol{y}_0 = \operatorname{DCT}(\boldsymbol{x}_0)$
            \STATE $\boldsymbol{c} = \operatorname{DCT}(\operatorname{Padding}(\boldsymbol{x}_{0}^{O}))$
		\STATE $t \sim \operatorname{Uniform}(\{1,2, \cdots, T\})$
		\STATE $\boldsymbol{\epsilon} \sim \mathcal{N}(\boldsymbol{0}, \boldsymbol{I})$
            \STATE $\boldsymbol{\theta}=\boldsymbol{\theta}-\nabla_{\boldsymbol{\theta}}\left\|\boldsymbol{\epsilon}-\boldsymbol{\epsilon}_{\boldsymbol{\theta}}\left(\sqrt{\bar{\alpha}_t} \mathbf{y}_0+\sqrt{1-\bar{\alpha}_t} \boldsymbol{\epsilon}, \boldsymbol{c}, t\right)\right\|^2$
            \ENDFOR
		\ENSURE  trained denoiser network $\epsilon_\theta$
	\end{algorithmic}
\end{algorithm}

\begin{algorithm}
	\renewcommand{\algorithmicrequire}{\textbf{Input:}}
	\renewcommand{\algorithmicensure}{\textbf{Output:}}
	\caption{Inference procedure of TransFusion}
	\label{alg2}
	\begin{algorithmic}[1]
            \REQUIRE observed motion sequence $\boldsymbol{x}^{O}$, diffusion steps $T$, the mask of the observation $\boldsymbol{M}$, trained denoiser $\epsilon_\theta$.
		\STATE $\boldsymbol{y}_{T} \sim \mathcal{N}(\boldsymbol{0}, \boldsymbol{I})$
            \STATE $\boldsymbol{y} = \boldsymbol{c} = \operatorname{DCT}(\operatorname{Padding}(\boldsymbol{x}^{O}))$
            \FOR{$t = T, T-1, \cdots,1$}
            \STATE $\boldsymbol{z} \sim \mathcal{N}(\boldsymbol{0}, \boldsymbol{I})$ if $t>1$, else $\boldsymbol{z}=\boldsymbol{0}$
            \STATE $\boldsymbol{y}^{O}_{t-1} = \sqrt{\bar\alpha_{t-1}} \boldsymbol{y} + \sqrt{1 - \bar\alpha_{t-1}} \boldsymbol{z}$
            \STATE $\boldsymbol{y}^{D}_{t-1}=\frac{1}{\sqrt{\alpha_t}}\left(\boldsymbol{y}_t-\frac{\beta_t}{\sqrt{1-\bar{\alpha}_t}} \boldsymbol{\epsilon}_{\boldsymbol{\theta}}\left(\boldsymbol{y}_t, \boldsymbol{c}, t\right)\right)+\sigma_t \boldsymbol{z}$
            \STATE $\boldsymbol{x}_{t-1}^{O} = \operatorname{IDCT}\left(\boldsymbol{y}_{t-1}^{O}\right)$
            \STATE $\boldsymbol{x}_{t-1}^{D} = \operatorname{IDCT}\left(\boldsymbol{y}_{t-1}^{D}\right)$
            \STATE $\boldsymbol{y}_{t-1} = \operatorname{DCT}\left(\boldsymbol{M} \odot \boldsymbol{x}_{t-1}^{O} + (\boldsymbol{1}-\boldsymbol{M}) \odot \boldsymbol{x}_{t-1}^{D}\right)$
            \ENDFOR
            \STATE $\boldsymbol{x} = \operatorname{IDCT}\left(\boldsymbol{y}_{0}\right)$
		\ENSURE  complete motion sequence $\boldsymbol{x}$
	\end{algorithmic}  
\end{algorithm}

\section{EXPERIMENTS}

\begin{table*}[htbp]
\caption{Quantitative results with best-of-many strategy on Human3.6M and HumanEva-I}
\begin{center}
\begin{threeparttable}
\begin{tabular}{cccccc|ccccc}
\toprule
            & \multicolumn{5}{c|}{Human3.6M}         & \multicolumn{5}{c}{HumanEva-I}        \\ \cmidrule{2-11}
Method      & APD    & ADE   & FDE   & MMADE & MMFDE & APD   & ADE   & FDE   & MMADE & MMFDE \\ \midrule
DeLiGAN \cite{c25}    & 6.509  & 0.483 & 0.534 & 0.520 & 0.545 & 2.177 & 0.306 & 0.322 & 0.385 & 0.371 \\
HP-GAN \cite{c26}     & 7.214  & 0.858 & 0.867 & 0.847 & 0.858 & 1.139 & 0.772 & 0.749 & 0.776 & 0.769 \\
Pose-Knows \cite{c27} & 6.723  & 0.461 & 0.560 & 0.522 & 0.569 & 2.308 & 0.269 & 0.296 & 0.384 & 0.375 \\
MT-VAE \cite{c28}     & 0.403  & 0.457 & 0.595 & 0.716 & 0.883 & 0.021 & 0.345 & 0.403 & 0.518 & 0.577 \\
DLow \cite{c29}       & 11.741 & 0.425 & 0.518 & 0.495 & 0.531 & 4.855 & 0.251 & 0.268 & 0.362 & 0.339 \\
DivSamp \cite{c30}    & \underline{15.310} & 0.370 & 0.485 & \underline{0.475} & \underline{0.516} & \textbf{6.109} & 0.220 & \underline{0.234} & \textbf{0.342} & \textbf{0.316} \\
BoM \cite{c32}        & 6.265  & 0.448 & 0.533 & 0.514 & 0.544 & 2.846 & 0.271 & 0.279 & 0.373 & 0.351 \\
DSF \cite{c33}        & 9.330  & 0.493 & 0.592 & 0.550 & 0.599 & 4.538 & 0.273 & 0.290 & 0.364 & 0.340 \\
MOJO \cite{c34}       & 12.579 & 0.412 & 0.514 & 0.497 & 0.538 & 4.181 & 0.234 & 0.244 & 0.369 & 0.347 \\
MultiObj \cite{c35}   & 14.240 & 0.414 & 0.516 & -     & -     & 5.786 & 0.228 & 0.236 & -     & -     \\
GSPS \cite{c31}       & 14.757 & 0.389 & 0.496 & 0.476 & 0.525 & 5.825 & 0.233 & 0.244 & \underline{0.343} & 0.331 \\
MotionDiff \cite{c40} & \textbf{15.353} & 0.411 & 0.509 & 0.508 & 0.536 & \underline{5.931} & 0.232 & 0.236 & 0.352 & \underline{0.320} \\
BeLFusion \cite{c41}  & 7.602  & 0.372 & \underline{0.474} & \textbf{0.473} & \textbf{0.507} & -     & -     & -     & -     & -     \\
HumanMAC \cite{c43}   & 6.301  & \underline{0.369} & 0.480 & 0.509 & 0.545 & $\cdots$    & \underline{0.209} & \textbf{0.223} & \textbf{0.342} & 0.335 \\ \midrule
TransFusion & 5.975       & \textbf{0.358}      & \textbf{0.468}      & 0.506      & 0.539      & 1.031   & \textbf{0.204}      & \underline{0.234}      & 0.408      & 0.427      \\ 
\bottomrule
\end{tabular}
\begin{tablenotes}
     \item[*] Bolded numbers indicate the best results, and numbers with underline represent the second best results. For all accuracy metrics, lower values are preferred. It is important to note that APD measures the difference among the 50 prediction results, and a larger APD does not necessarily indicate better performance. The symbol `-' indicates that certain results are not reported in the baselines, and `$\cdots$' denotes that the result reported in the baseline has some issue.
\end{tablenotes}
\label{best-of-many}
\end{threeparttable}
\end{center}
\vspace{-0.1 in}
\end{table*}

\begin{table*}[htbp]
\caption{Quantitative results with median-of-many and worst-of-many strategy on Human3.6M}
\centering
\begin{threeparttable}
\begin{tabular}{ccccc|cccc}
\toprule
            & \multicolumn{4}{c|}{Median-of-many} & \multicolumn{4}{c}{Worst-of-many} \\ \cmidrule{2-9} 
Model       & ADE-M    & FDE-M    & MMADE-M   & MMFDE-M   & ADE-W   & FDE-W   & MMADE-W   & MMFDE-W   \\ \midrule
DLow \cite{c29}    & 0.896       & 1.285       & 0.948        & 1.290        & 1.763      & 2.655      & 1.804        & 2.657        \\
DivSamp \cite{c30}    & 0.924       & 1.344       & 1.001        & 1.359       & 2.497      & 3.263      & 2.530        & 3.267        \\
GSPS \cite{c31}    & 1.014       & 1.375       & 1.066        & 1.383        & 2.458      & 2.964      & 2.480        & 2.964        \\
BeLFusion \cite{c41}    & 0.673       & 0.976       & 0.767        & 1.009        & 1.355      & 2.038      & 1.418        & 2.046        \\
HumanMAC \cite{c43}    & \underline{0.585}       & \underline{0.911}       & \underline{0.736}        & \underline{0.977}        & \underline{1.085}      & \underline{1.843}      & \underline{1.205}        & \underline{1.877}        \\ \midrule
TransFusion & \textbf{0.575}       & \textbf{0.898}       & \textbf{0.729}        & \textbf{0.967}        & \textbf{1.063}      & \textbf{1.758}      & \textbf{1.179}        & \textbf{1.791}        \\ \bottomrule
\end{tabular}
\begin{tablenotes}
     \item[*] Quantitative results of baselines are calculated from pretrained models.
\end{tablenotes}
\label{median}
\end{threeparttable}
\vspace{-0.15 in}
\end{table*}

    \begin{figure}
      \centering
      \includegraphics[width=1.0\columnwidth]{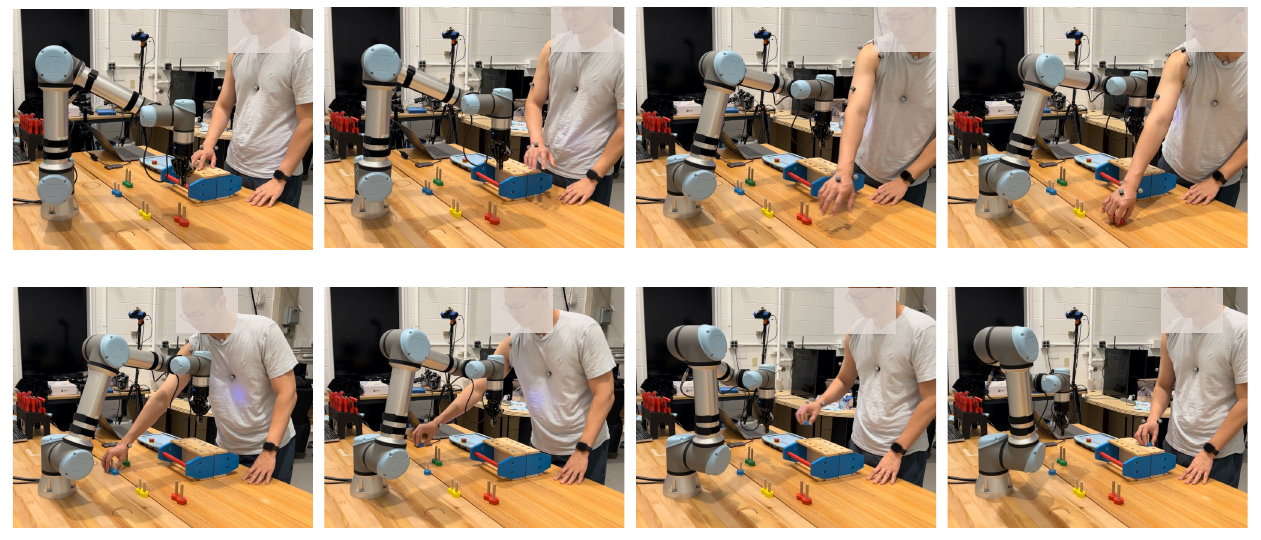}
      \vspace{-0.1in}
      \caption{The HRC reaching motion dataset scenario, depicting human-robot collaboration in collecting screws on the table. The top row illustrates the human reaching out to pick up a screw, while the bottom row showcases the human bringing a screw back. Motion data is recorded using markers attached to the human body.}
      \label{hrc_scenario}
      \vspace{-0.1 in}
   \end{figure}

\subsection{Experimental Setup}

\textbf{Datasets.} We evaluate the performance of our proposed human motion prediction model, TransFusion, on two benchmark datasets: Human3.6M \cite{c7} and HumanEva-I \cite{c8}. Human3.6M is the largest benchmark dataset for human motion prediction and analysis, comprising 3.6 million frames of human poses recorded at 50 Hz. The dataset consists of 17 daily actions, such as walking, smoking, discussion, and taking photos, performed by 11 actors. To ensure a fair comparison with other works, we adopt the widely-used setting proposed by \cite{c29}. Specifically, we represent the human pose using a subset of 17 joints, rather than the original 32-joint human skeleton. The model is trained on 5 subjects (S1, S5, S6, S7, and S8) and tested on 2 subjects (S9 and S11). For prediction, we utilize 25 frames (0.5 seconds) as the observation to forecast the following 100 frames (2 seconds). On the other hand, HumanEva-I is a relatively smaller dataset, where 4 subjects perform 6 common actions, such as walking and jogging. Compared to Human3.6m, HumanEva-I exhibits less variation in motion and is recorded at a higher frequency of 60Hz. In line with prior works, we represent the human skeleton using 15 joints and follow the official train/test split provided in the original dataset. For HumanEva-I, the prediction horizon is set to 60 frames (1 second) given an observation of 15 frames (0.25 seconds).

We also conduct performance evaluations on a HRC reaching motion dataset \cite{c5-1}, where the human and robot collaborate in collecting screws from various locations on a table, as depicted in Fig. \ref{hrc_scenario}. The dataset contains motion data of both the human and robot agents. The human agent is represented as a 5-joint skeleton, consisting of the xiphoid process, the incisura jugularis, the shoulder, the elbow, and the wrist, and the robot manipulator is represented as an 8-joint skeleton. Mocap system is employed to record the human motion at a frequency of 50 Hz. For training purposes, we utilize 400 data samples, and 63 data samples are reserved for testing. The prediction period is set to 60 frames (equivalent to 1.2 seconds), given an observation window of 15 frames (equivalent to 0.3 seconds).

\textbf{Evaluation metrics.} We adopt the commonly-used pipeline proposed in \cite{c29} and use five metrics to evaluate our model's performance: (1) Average Pairwise Distance (APD): This metric computes the average L2 distance between all pairs of motion samples, serving as a measure of diversity within the predicted future motions. (2) Average Displacement Error (ADE): ADE calculates the smallest average L2 distance over all time steps between the ground truth and predicted samples, evaluating the prediction accuracy. (3) Final Displacement Error (FDE): FDE measures the smallest L2 distance in the last time frame between the prediction results and ground truth, also evaluating theprediction accuracy. (4) Multi-Modal ADE (MMADE): This metric is the multi-modal version of ADE, assessing the model's ability to capture the multi-modality nature of human motion. Multi-modal ground truth future motions are obtained by grouping similar observations. (5) Multi-Moddal FDE (MMFDE): Similar to MMADE, MMFDE is the multi-modal version of FDE, examining the model's capacity to predict multiple plausible outcomes.

While these metrics provide valuable insights, we argue that relying solely on them may not be sufficient. Prior works often prioritize increasing the APD to enhance diversity, sometimes leading to unrealistic and implausible predictions. Moreover, the accuracy evaluation of previous works is based on the best-of-many strategy, where only the closest sample to the ground truth is considered, potentially overlooking predictions that deviate significantly from reality. This approach might not be suitable for real-world tasks, such as human-robot collaboration.

In our study, we aim to generate not just one good sample close to the ground truth but as many good predictions as possible while maintaining a certain degree of diversity. To achieve this, we introduce additional strategies: worst-of-many and median-of-many, along with their corresponding metrics: (6) ADE-W, (7) FDE-W, (8) MMADE-W, (9) MMFDE-W for worst-of-many evaluation, and similarly, (10) ADE-M, (11) FDE-M, (12) MMADE-M, (13) MMFDE-M for median-of-many evaluation. By incorporating these strategies, we offer a more comprehensive evaluation of our model's performance, considering both accuracy and diversity in the predictions.

\textbf{Baselines.} To assess the effectiveness of our model, we conduct a comparative evaluation with several state-of-the-art works, which include DeLiGAN \cite{c25}, HP-GAN \cite{c26}, Pose-Knows \cite{c27}, MT-VAE \cite{c28}, DLow \cite{c29}, DivSamp \cite{c30}, BoM \cite{c32}, DSF \cite{c33}, MOJO \cite{c34}, MultiObj \cite{c35}, GSPS \cite{c31}, MotionDiff \cite{c40}, BeLFusion \cite{c41}, and HumanMAC \cite{c43}. Notably, we exclude TCD \cite{c42} from the comparison as it adopts a two-stage prediction strategy, which can be readily combined with other prediction models. Therefore, considering TCD as a baseline in this context would not be fair.

    \begin{figure*}
      \centering
      \includegraphics[width=2\columnwidth]{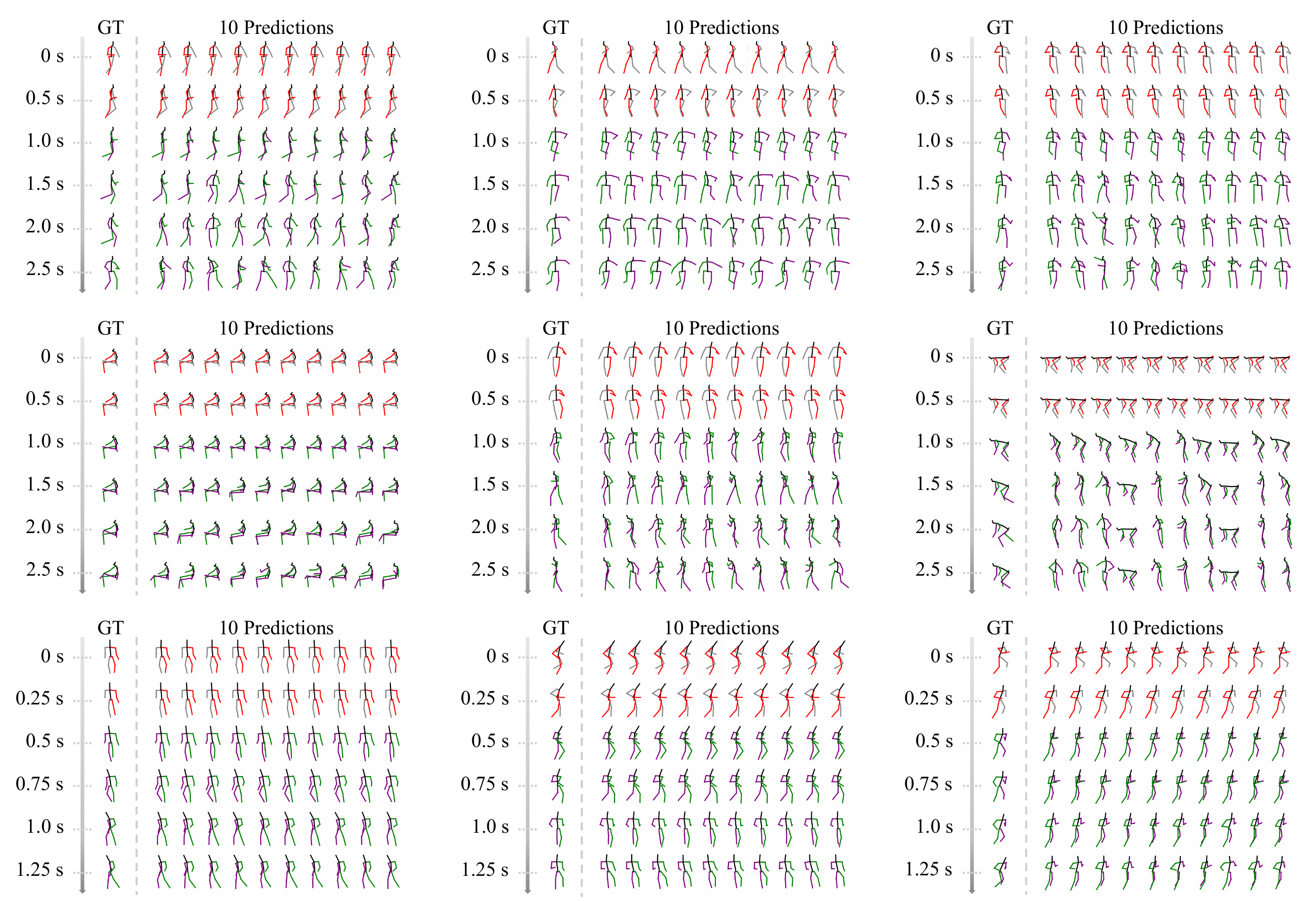}
      \caption{Visualization results. The first two rows display the results from Human3.6M, with each example corresponding to different scenarios: 'Walking', 'Walking Together', 'Walking Dog', 'Sitting on Chair', 'Smoking', and 'Making Purchases', respectively. The last row presents the results from HumanEva-I, with the labels being 'Walking', 'Jogging', and 'Boxing'. For each sample, the first column represents the ground truth, while the following 10 columns depict the prediction results. The observed context for each motion is represented by red and black skeletons, while the future motion is indicated by green and purple skeletons.}
      \label{vis}
   \end{figure*}

\textbf{Implementation details.} We train TransFusion with 1,000 diffusion steps with the variance schedule based on the cosine scheduler \cite{c50}. During training, we sample 50,000 data from the training set in each epoch for both benchmark datasets. We train the model for 1,500 epochs on Human3.6M with a batch size of 64 and 100 epochs on HumanEva-I with the same batch size. We also adopt classifier-free guidance \cite{c51} during training, where there is a 0.2 probability of disregarding historical conditions. The learning rate is initialized to $3 \times 10^{-4}$ and is decayed by a ratio of 0.8 every 100 epochs. For the noise prediction network, we use 9-layer SE-Transformer blocks for Human3.6M and 5-layer SE-Transformer blocks for HumanEva-I. Furthermore, we use the first 20 rows of DCT coefficients for both Human3.6M and HumanEva-I. Following common practice, we set the dimension of the hidden state to 512 for both benchmark datasets. To expedite the inference process, we leverage a 100-step DDIM \cite{c48} and generate 50 predictions for each single observation. All experiments are conducted using PyTorch \cite{c52} and a single NVIDIA Tesla V100 GPU. Adam \cite{c53} is used as the optimizer for all experiments.

\begin{table}[htbp]
\caption{Complexity comparison}
\begin{center}
\begin{threeparttable}
\begin{tabular}{ccccc}
\toprule
              &          & \multicolumn{3}{c}{Human3.6M (Best-of-many)} \\ \cmidrule{3-5} 
Model         & \#Params & APD      & ADE      & FDE     \\ \midrule
HumanMAC-8 \cite{c43}      & 28.40M         & 6.301         & 0.369         & 0.480        \\
TransFusion-7 & \textbf{15.52M}         & \textbf{6.537}         & \underline{0.363}         & \underline{0.471}        \\
TransFusion-9 & \underline{19.73M}         & \underline{5.975}         & \textbf{0.358}         & \textbf{0.468}        \\ \bottomrule
\end{tabular}
\begin{tablenotes}
     \item[*] The number following the model name indicates the number of layers in the noise prediction network.
\end{tablenotes}
\end{threeparttable}
\end{center}
\label{params}
\vspace{-0.15 in}
\end{table}

\subsection{Comparison with the State-of-the-Arts}

We first follow the best-of-many strategy to compare TransFusion with existing works. The quantitative results on both benchmark datasets are presented in Table \ref{best-of-many}. For Human3.6M, our model outperforms all baselines, achieving state-of-the-art performance on ADE and FDE metrics. On HumanEva-I, TransFusion achieves the best result on ADE and the second-best result on FDE. These findings indicate that our model generates predictions that are closer to the ground truth compared to other works. While the APD results from our model are not as favorable as some prior works, we have already established that the higher APD does not necessarily imply better predictions, and sometimes the situation can be opposite. Excessive diversity can lead to predictions that significantly deviate from reality, resulting in unrealistic and overly conservative outcome. As a result, such methods may hinder efficiency in downstream tasks or even fail to meet the requirements of certain applications, such as motion planning for robotic manipulators in human-robot collaboration. Therefore, we prioritize a balanced approach and do not solely focus on increasing the APD in this work.

Since the commonly-used strategy only evaluates the quality of the closest prediction, we also assess the overall prediction quality by providing results following the median-of-many and worst-of-many strategies in Table \ref{median}. As shown in the table, TransFusion surpasses all other baselines accross all accuracy metrics, highlighting its superior overall prediction performance. We further illustrate the quality of the motion predicted by our model through visualization in Fig. \ref{vis}.

We also observe that HumanMAC \cite{c43} achieves slightly inferior performance compared to ours. To delve deeper, we compare the total number of parameters of both models in Table \ref{params}. Remarkably, TransFusion achieves better results than HumanMAC while utilizing only 54.6\% of the parameters employed by the latter. This efficiency gain is attributed to the fact that our model incorporates conditions without relying on any additional modules, unlike other existing works.

\begin{figure}
      \centering
      \includegraphics[width=1.0\columnwidth]{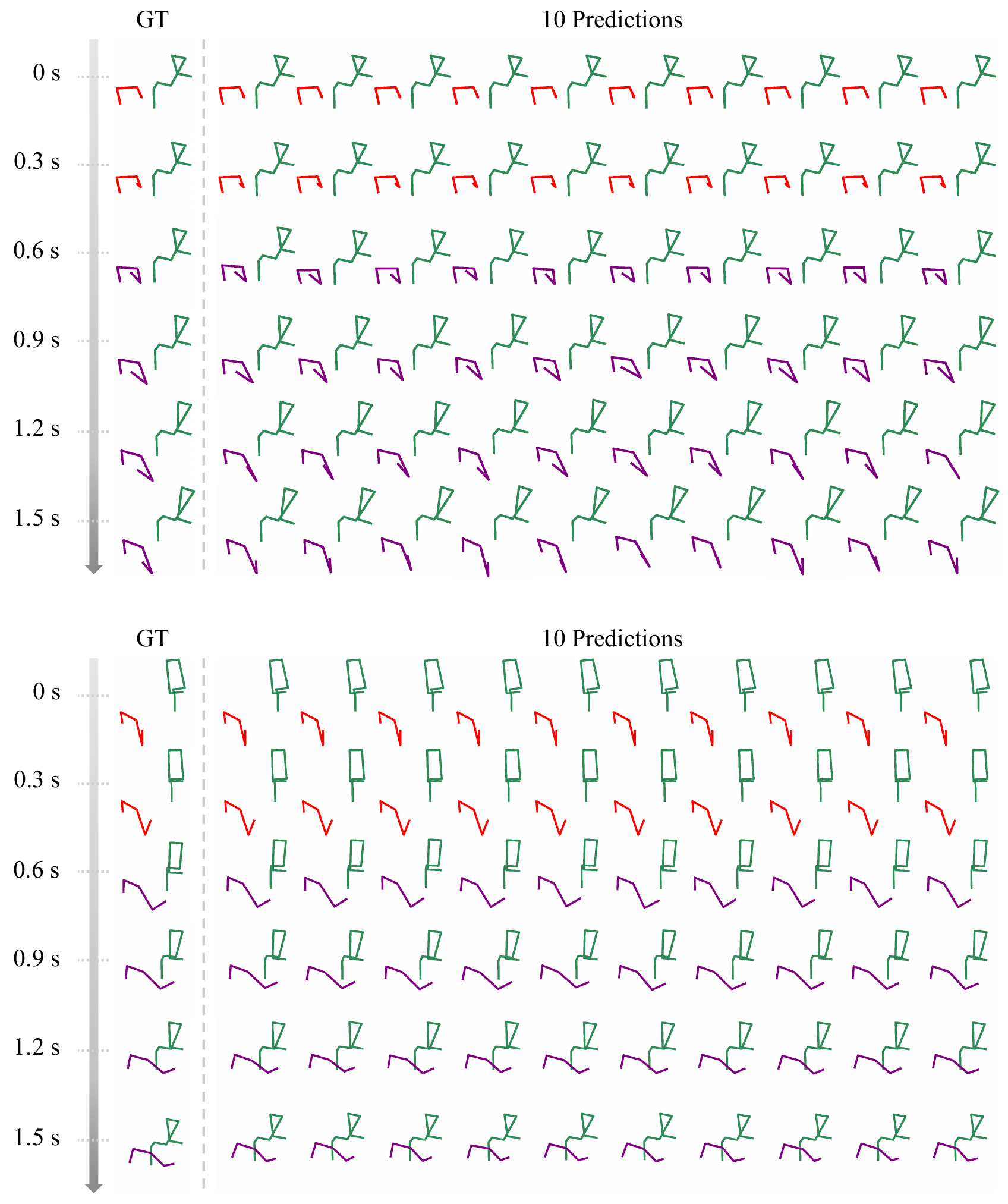}
      \vspace{-0.1in}
      \caption{Visualization results on the HRC reaching motion dataset. The green skeletons depict the robot manipulator, the red skeletons illustrate the historical motion of the human's upper body and right arm, and the purple skeletons correspond to the predicted future motion of the human. The top case shows the motion of the human bringing a screw back, while the bottom case displays the motion of the human reaching out to pick up a screw.}
      \label{hrc_results}
   \end{figure}

Moreover, we adopt the best-of-50 strategy to evaluate our model on the HRC reaching motion dataset. The results demonstrate an APD equal to 0.727, an ADE equal to 0.035, and a FDE equal to 0.047. These metrics reinforce the applicability of our model to HRC scenarios while accounting for uncertainty. The visualization of the results can be found Fig. \ref{hrc_results}.

\subsection{Ablation Study}

We conduct comprehensive ablation studies to explore different design choices of TransFusion, including the skip connections, the utilization of SE block, the number of rows ($L$) we used in DCT/IDCT, and the number of layers in the noise prediction network. The following evaluations are based on the best-of-many strategy. 

\begin{table}[htbp]
\begin{center}
\caption{Quantitative results of the ablation study on the skip connections}
\begin{tabular}{cccc|ccc}
\toprule
   & \multicolumn{3}{c|}{Human3.6M} & \multicolumn{3}{c}{HumanEva-I} \\ \cmidrule{2-7} 
Model  & APD      & ADE      & FDE      & APD      & ADE      & FDE      \\ \midrule
Concat  & 5.975         & \textbf{0.358}         & \textbf{0.468}         & \textbf{1.031}         & \textbf{0.204}         & \textbf{0.234}         \\
Add & 7.224         & 0.404        & 0.508         & 0.996         & 0.206         & 0.235         \\
w/o skip & \textbf{7.447}         & 0.411         & 0.511         & 0.945        & 0.205         & 0.237         \\ \bottomrule
\end{tabular}
\label{ablation-skip}
\vspace{-0.15 in}
\end{center}
\end{table}

\begin{table}[htbp]
\begin{center}
\caption{Quantitative results of the ablation study on SE block}
\begin{tabular}{cccc|ccc}
\toprule
   & \multicolumn{3}{c|}{Human3.6M} & \multicolumn{3}{c}{HumanEva-I} \\ \cmidrule{2-7} 
Model  & APD      & ADE      & FDE      & APD      & ADE      & FDE      \\ \midrule
w SE  & 5.975         & \textbf{0.358}         & \textbf{0.468}         & \textbf{1.031}         & \textbf{0.204}         & \textbf{0.234}         \\
w/o SE & \textbf{6.071}         & 0.361         & 0.472         & 1.004         & 0.210         & 0.237         \\ \bottomrule
\end{tabular}
\label{ablation-se}
\vspace{-0.15 in}
\end{center}
\end{table}

\begin{table}[htbp]
\begin{center}
\caption{Quantitative results of the ablation study on L}
\begin{tabular}{cccc|ccc}
\toprule
   & \multicolumn{3}{c|}{Human3.6M} & \multicolumn{3}{c}{HumanEva-I} \\ \cmidrule{2-7} 
L  & APD      & ADE      & FDE      & APD      & ADE      & FDE      \\ \midrule
5  & 5.738         & 0.381         & 0.493         & 0.797         & 0.227         & 0.284         \\
10 & 5.790         & 0.364         & 0.473         & 0.931         & \textbf{0.204}         & 0.244         \\
20 & 5.975         & \textbf{0.358}         & \textbf{0.468}         &  1.031        & \textbf{0.204}         & 0.234         \\
30 & \textbf{6.018}         & 0.360         & 0.470         & \textbf{1.136}         & \textbf{0.204}         & \textbf{0.229}         \\ \bottomrule
\end{tabular}
\label{ablation-L}
\end{center}
\vspace{-0.15 in}
\end{table}

\begin{table}[htbp]
\begin{center}
\caption{Quantitative results of the ablation study on \#Layers}
\begin{tabular}{cccc|ccc}
\toprule
   & \multicolumn{3}{c|}{Human3.6M} & \multicolumn{3}{c}{HumanEva-I} \\ \cmidrule{2-7} 
\#Layers  & APD      & ADE      & FDE      & APD      & ADE      & FDE      \\ \midrule
3  & \textbf{8.172}         & 0.416         & 0.533         & \textbf{1.288}         & 0.210         & 0.238         \\
5 & 6.494         & 0.374         & 0.485         & 1.031         & 0.204         & \textbf{0.234}         \\
7 & 6.537         & 0.363         & 0.471         & 0.923         & \textbf{0.203}         & 0.236         \\
9 & 5.975         & \textbf{0.358}         & \textbf{0.468}         & 0.872         & 0.206         & 0.239         \\
11 & 5.722         & 0.360         & 0.472         & 0.846         & 0.205         & 0.243         \\ \bottomrule
\end{tabular}
\label{ablation-layers}
\end{center}
\vspace{-0.15 in}
\end{table}

We begin by evaluating the effectiveness of long skip connections in our network. As shown in Table \ref{ablation-skip}, it is evident that the inclusion of skip connections enhances the model's performance, as both the concatenation and addition-based skip connections outperform the model without skip connections. Moreover, the model with concatenation-based skip connections demonstrates superior performance compared to the one with addition-based skip connections. We argue that the addition-based method of simply adding the branches together, without employing a linear projection, does not significantly benefit the model learning process. This is because in the addition-based method, the deeper layers already have a direct path from shallower layers due to the presence of residual connections in SE-Transformer blocks. As a result, the concatenation-based skip connections offer more advantages, leading to improved model performance.

As previously mentioned, we add an SE block in the Transformer encoder to optimize learning. The results shown in Table \ref{ablation-se} indicate that the addition of the SE module indeed enhances the model's performance, as evidenced by improved ADE and FDE results on both datasets.

Additionally, we investigate the impact of the dimensionality of the problem by using only the first $L$ rows of DCT basis. Smaller values of $L$ may result in the loss of important information, while larger values may add computational burden to the model. Therefore, we assess the influence of $L$ on the model's performance, and the results are provided in Table \ref{ablation-L}. For Human3.6M, the best ADE and FDE metrics are achieved when $L=20$, whereas for HumanEva-I, $L=30$ performs the best. However, since $L=20$ achieves the same ADE as $L=30$ on HumanEva-I, we opt to set $L$ equal to 20 for both datasets.

Table \ref{ablation-layers} presents the results of experiments with different number of layers. For Human3.6M, we use 9 layers in our model as it yields the best performance in both ADE and FDE. On the other hand, for HumanEva-I, a 5-layer network shows the best FDE, while a 7-layer network performs best in terms of ADE. Considering efficiency, we finally choose the 5-layer network for HumanEva-I.

\section{CONCLUSIONS}

This paper presents a practical and effective diffusion-based human motion prediction method, leveraging the Transformer as the backbone with long skip connections between shallow and deep layers. We design the model in the frequency domain, utilizing the DCT operation. To condition the predictions on historical information, we treat the conditions as a token, avoiding the use of any additional module like cross-attention and adaptive normalization. The extensive experimental studies demonstrate that our model achieves state-of-the-art prediction results in terms of accuracy. In contrast to prior works often prioritize diversity and produce unrealistic future motions, TransFusion stands out by offering superior overall prediction quality while still maintaining a certain degree of diversity. The model's ability to strike a balance between accuracy and diversity makes it a promising solution for human motion prediction tasks.

\end{document}